# An MLP based Approach for Recognition of Handwritten 'Bangla' Numerals

Subhadip Basu [+] [ξ], Nibaran Das*, Ram Sarkar*,
Mahantapas Kundu*, Mita Nasipuri*, Dipak Kumar Basu*

[+] Computer Sc. & Engg. Dept., MCKV Institute of Engineering,
Liluah, Howrah-711204, India.
*Computer Sc. & Engg. Dept., Jadavpur University,
Kolkata-700032, India.
[ξ] Corresponding Author; e-mail: {subhadip8@yahoo.com}

**Abstract.** The work presented here involves the design of a Multi Layer Perceptron (MLP) based pattern classifier for recognition of handwritten Bangla digits using a 76 element feature vector. Bangla is the second most popular script and language in the Indian subcontinent and the fifth most popular language in the world. The feature set developed for representing handwritten Bangla numerals here includes 24 shadow features, 16 centroid features and 36 longest-run features. On experimentation with a database of 6000 samples, the technique yields an average recognition rate of 96.67% evaluated after three-fold cross validation of results. It is useful for applications related to OCR of handwritten Bangla Digit and can also be extended to include OCR of handwritten characters of Bangla alphabet.

## 1 Introduction

Optical Character Recognition (OCR) systems appear to ease the interface between man and machine and help in office automation with huge saving of time and human effort. The OCR system allows desired manipulation of the scanned text as the output is coded with ASCII or some other character code of such system prepared from the paper based input text. Success of the commercially available OCR system is yet to be extended to handwritten text. It is mainly due to the fact that numerous variations in writing styles of individuals make recognition of handwritten characters difficult. For a specific language based on some alphabet and numerals, OCR techniques are either aimed at printed text or handwritten text. The present work is aimed at the latter.

Broadly speaking, OCR systems ease the barrier of the keyboard interface between man and machine to a great extent, and help in office automation with huge saving of time and human effort. The systems have potential applications in extracting data from filled in forms, interpreting handwritten addresses from postal documents for automatic routing, automatic reading of bank cheques etc.

Past work on OCR of handwritten alphabet and numerals has been mostly found to concentrate on Roman script [1] related to English and some European languages, and scripts related to Asian languages like Chinese [2], Korean, and Japanese. Among Indian scripts, Devnagri, Tamil, Oriya and Bangla have started to receive attention for OCR related research in the recent years.

Majority of the past work e.g. [1], [6], [7], [8], related to offline handwritten character recognition, concentrated on the analysis of English scripts. One of the most notable early work on handwritten English cursive word recognition is that of Bozinovic and Srihari [1], in which a word image is transformed through a hierarchy of representation levels: points, contours, features, letters, and words. They have used a bottom-up technique to generate an unique feature representation from the input image using statistical dependence between letters and features.

Apart from the work on the recognition of English script, a lot of significant work on the recognition of different Asian languages have also been done. Wong et. al. [2] worked on the analysis of Chinese script. They have used a character template codebook to match an unknown character. A. Amin in [3] presents a comparative study of various aspects of the machine recognition processes and research possibilities of off-line Arabic characters.

Visibly, a variety of research work have been done on OCR over the years. But certain popular languages have not yet received sufficient attention in this regard Bangla is one such language. It is the fifth most popular language in the world [4], a wide-spoken language in India and the national language of Bangladesh as well. About 200 million people of Eastern India and Bangladesh use this language for communication [4]. Moreover, Assamese and Manipuri, two popular languages in India, are also written in Bangla script. Two of the important research contributions relating to OCR of Bangla characters involve a *multistage approach* developed by Rahman et al. [9] and an MLP classifier developed by Bhowmik et al. [10]. The major features used for the multistage approach include *Matra*, upper part of the character, disjoint section of the character, vertical line and double vertical line. And, for the MLP classifier, the feature set is constructed from the stroke features of characters. S. Basu et. al. [5], have recently developed a two-pass approach for offline recognition of handwritten Bangla numerals. The work presents an effective technique for improvement of recognition performances with multiple classifiers.

**Fig.1.** The decimal digit set of Bangla script

In the light of above facts, the present work concentrates on the development of an MLP based pattern classifier for recognition of handwritten Bangla digits with a feature set of 76 features. There is enough scope for extension of this work to include handwritten characters of Bangla alphabet also. For experimentation on the present technique, we have used the standard handwritten Bangla digit set provided by the Indian Statistical Institute. Three fold cross validation of results are performed to find an average recognition performance of the MLP classifier designed for the present work. The results show a sharp increase of average recognition performance with the new feature set used here in comparison to one we previously used for design of the first pass classifier of the two pass approach [11]. In respect to the need of a wide cross section of world population using *Bangla* as a script or language, OCR of handwritten *Bangla* digits has a high value of commercial importance in the present time. It is still an active area of research in Pattern Recognition and Image Processing. The work presented here not only targets the development of a suitable feature set for representation of handwritten *Bangla* digits in the feature space but also the development of an effective classification technique for dealing with the same. All this has motivated the idea of the present work. Fig. 1 shows the typical digit patterns of the decimal digits of *Bangla* script.

## 2   The Feature Sets

Choice of suitable features for pattern classes, as mentioned before, is a domain specific design Issue. In the present work, two different feature sets are designed for classification of handwritten Bangla digit patterns. The feature sets are denoted by Feature Set #1 and Feature Set #2. The features constituting these two sets are so selected that their values remain close to each other for the patterns of the same class and differ appreciably for the patterns of different classes. The feature sets should be capable of supplying complementary information about the digit patterns, at least to some extent. For extraction of features from the digit images, the same are first enclosed within minimal bounding boxes and then scaled to 32X32 pixels sizes. The scaled images, which are defined with gray scale pixel values, are finally converted to binary images through thresholding.

**Feature Set # 1**

It consists of 40 features in all. These features are formed with 24 *shadow features* and 16 *octant centroid features* as described below:

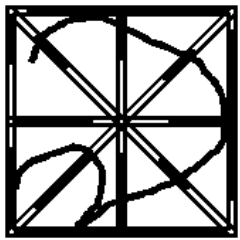

**Fig. 2.** An Illustration of the 24 Shadow features

**Shadow Features.** Shadows features are computed by considering the lengths of projections of the digit images, as shown in Fig. 2, on the four sides and eight octant dividing sides of the minimal bounding boxes enclosing the same. Considering the lengths of projections on three sides of each such octant, 24 shadow features are extracted from each digit image, which is divided into eight octants inside the minimal box. Each value of the shadow feature so computed is to be normalized by dividing it with the maximum possible length of the projections on the respective side.

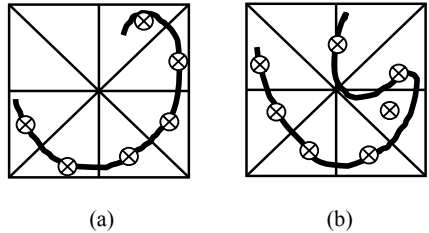

**Fig. 3 (a-b)**. An Illustration of the 16 Centroid features

**Octant Centroid Features.** Coordinates of centroids of black pixels in all the 8 octants of a digit image are considered to add 16 features in all to the feature set. Fig. 3(a-b) shows approximate locations of all such centroids on two different digit images. It is noteworthy how these features can be of help to distinguish the two images.

**Feature Set # 2**

It consists of 36 longest run features in all. It is described below.

**Longest Run Features.** For computing longest-run features from a character image, the minimal bounding box enclosing the image is divided into 9 overlapping rectangular regions. Coordinates (r, c) of top left corners of all these regions, in terms of the row number r and the column number c, are given as follows: {(r, c) | r=0, h/4, 2h/4 and c=0, w/4, 2w/4}, where h and w denote the height and the width of the minimal bounding box respectively. In each such rectangular region, 4 longest-run features are computed row wise, column wise and along two of its major diagonals.

The row wise longest-run feature is computed by considering the sum of the lengths of the longest bars that fit consecutive black pixels along each of all the rows of a rectangular region, as illustrated in Fig. 4(a-b). In fitting a bar with a number of consecutive black pixels within a rectangular region, the bar may be extended beyond the boundary of the region if the same is continued there. The three other longest-run features are computed in the same way but along the column wise and two major diagonal wise directions within the rectangle separately. Thus, in all, 9x4=36 longest-run features are computed from each character image. Each of these feature values is to be normalized by dividing the same with h x w. The product, h x w, represents the sum of the lengths of the bars, that fit consecutive black pixels individually in each of the four directions within the minimal square completely filled with black pixels.

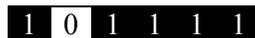

|   |   |   |   |   |   |
|---|---|---|---|---|---|
| 1 | 0 | 0 | 1 | 1 | 0 |
| 1 | 0 | 0 | 1 | 1 | 0 |
| 1 | 0 | 0 | 0 | 1 | 0 |
| 0 | 1 | 0 | 0 | 1 | 0 |
| 0 | 0 | 1 | 1 | 0 | 0 |

(a)

| | | | | | | Length of the Longest Bar |
|---|---|---|---|---|---|---|
| 1 | 0 | 4 | 4 | 4 | 4 | 4 |
| 1 | 0 | 0 | 2 | 2 | 0 | 2 |
| 1 | 0 | 0 | 2 | 2 | 0 | 2 |
| 1 | 0 | 0 | 0 | 1 | 0 | 1 |
| 0 | 1 | 0 | 0 | 1 | 0 | 1 |
| 0 | 0 | 2 | 2 | 0 | 0 | 2 |
| | | | | | Sum = | 12 |

(b)

**Fig 4.** An illustration for computation of the row wise longest–run feature.
   (a) The portion of a binary image enclosed within a rectangular region.
   (b) Every pixel position in each row of the image is marked with the length of the longest bar that fits consecutive black pixels along the same row.

## 3   The MLP Classifier

In the present work, an MLP classifier is employed for recognition of unknown digit patterns using Feature Set # 1 and Feature Set # 2. The MLP is a special kind of Artificial Neural Network (ANN). ANNs are developed to replicate *learning* and *generalization* abilities of human's behaviour with an attempt to model the functions of *biological neural networks* of the human brain.

      Architecturally, an MLP is a feed-forward layered network of *artificial neurons*. Each artificial neuron in the MLP computes a *sigmoid function* of the weighted sum of all its inputs. An MLP consists of one *input layer*, one *output layer* and a number of *hidden* or intermediate *layers*, as shown in Fig 5. The output from every neuron in a layer of the MLP is connected to all inputs of each neuron in the immediate next layer of the same. Neurons in the input layer of the MLP are all basically dummy neurons as they are used simply to pass on the input to the next layer just by computing an identity function each.

The numbers of neurons in the input and the output layers of an MLP are chosen depending on the problem to be solved. The number of neurons in other layers and the number of layers in the MLP are all determined by a trial and error method at the time of its *training*. An ANN requires training to learn an unknown input-output relationship to solve a problem.

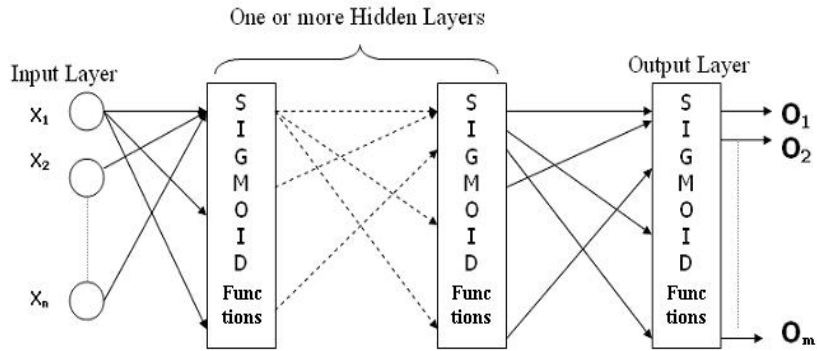

**Fig. 5.** A block diagram of an MLP shown as a feed forward neural network.

Depending on the models of ANNs, training is performed either under supervision of some teacher (i.e., with labeled data of known input-output responses) or without supervision. The MLP to be used for the present work requires supervised training. During training of an MLP *weights* or strengths of neuron-to-neuron connections, also called *synapses*, are iteratively tuned so that it can respond appropriately to all training data and also to other data, not considered at the time of training. Learning and generalization abilities of an ANN is determined on the basis of how best it can respond under these two respective situations.

The MLP classifier designed for the present work is trained with the Back Propagation (BP) algorithm. It minimizes the *sum of the squared errors* for the training samples by conducting a *gradient descent* search in the *weight space*. The number neurons in a hidden layer in the same are also adjusted during its training.

The problem of *pattern classification* involves two successive transformations as follows:

$$M \rightarrow F \rightarrow D$$

Where, M, F and D stand for the measurement space, the feature space and the decision space respectively. Once a feature set is fixed up, it is left with the design of a mapping ($\delta$) as follows:

$$\delta: F \rightarrow D$$

ANNs with their learning and generalization abilities can approximate a general class of functions given below.

$$f: \mathbb{R}^n \to \mathbb{R}$$

Pattern classification with ANNs requires approximating δ as a *discrete valued function* shown below.

$$\delta: \mathbb{R}^n \to \{1, 2, ..m\}$$

where, n and m denotes the number of features and the number of pattern classes respectively. So an ANN based pattern classifier requires n number of neurons in the input layer and m number of neurons in the output layer. Conventionally 1-out-of-m representation is used for its output.

## 4 Results and Discussion

MLP classifiers require training with labeled patterns of samples before they start operation as pattern classifiers. Training is followed by testing to check whether an MLP classifier can generalize from test data on the basis of what it has learnt from training data. For preparation of the training and the test sets of samples, a database of 6,000 Bangla digit samples is formed by collecting optically scanned handwritten digit samples of 10 digit symbols from each of 600 people of different age groups and sexes. A *training set* of 4000 samples and a *test set* of 2000 samples are then formed. Three such pairs of the training and the test sets are formed in all with the original database for cross validation of results at the time of experimentation. All the samples are scaled to 32x32 pixel images first and then converted to binary images through thresholding.

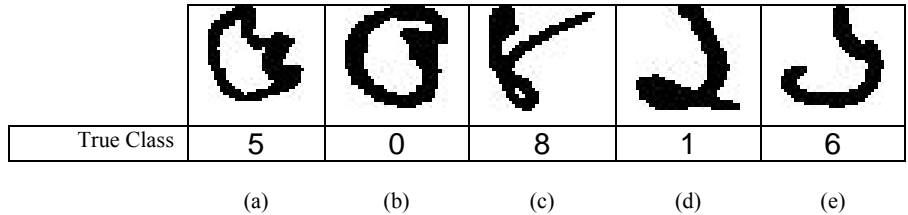

**Fig. 6 (a-e).** Some test Samples classified truly through the MLP classifier.

For the present work, a single layer MLP, i.e., an MLP with one hidden layer is chosen for each pair of 4000 training samples and 2000 test samples. This is mainly to keep the computational requirement of the same low without affecting its function approximation capability. To design an MLP for classification of handwritten alphabetic characters, several runs of BP algorithm with *learning rate* (η) = 0.8 and *momentum term* (α)=0.7 are executed for different numbers of neurons in its hidden layer. Recognition performances of the MLP on the test samples of the three sets observed from this experimentation are given in Table1.

**Table 1.** Recognition performances of the MLP with different numbers of neurons in the hidden layers

| No of Hidden neurons | Percentage recognition rate of the MLP on test samples | | |
|:---:|:---:|:---:|:---:|
| | *Set#1* | *Set#2* | *Set#3* |
| 25 | 95.5 | 96.1 | 95.65 |
| 30 | 96.1 | 96.05 | 96 |
| 35 | 95.85 | 95.9 | 96.15 |
| 40 | 96.2 | 96.15 | 96.65 |
| 45 | 96.3 | 96.05 | 96.25 |
| 50 | 96.05 | 95.95 | 96.6 |
| 55 | 95.95 | 96.1 | 96.7 |
| 60 | 96.1 | 96 | 96.8 |
| **65** | **96.65** | **96.7** | **96.65** |
| 70 | 96.15 | 96.1 | 96.6 |

Curves showing variation of the Recognition performance of the MLP, for the test samples of the three sets, with increase in the number of neurons in its hidden layer are plotted in Fig. 7, Fig. 8 and Fig. 9 from the Table 1 respectively. It is required to fix up the number of neurons in the hidden layer of MLP so that it can show the optimal recognition performance on the test set.

Recognition performances of the MLP for the Set#2, as observed from the curve shown in Fig. 8, initially rise as the number of neurons in the hidden layer is increased and falls after the same crosses some limiting value. It reflects the fact that for some fixed training and test sets, learning and generalization abilities of the MLP improve as the number of neurons in its hidden layer as increases up to certain limiting value and any further increase in the number of neurons in the hidden layer thereafter degrades the abilities. It is called the *over-fitting* problem.

The optimal recognition performance of the MLP is observed at a point, on the curve of Fig. 7, where the number of neurons in its hidden layer is set to 65 in Set#1. Similarly, for Set#2 the optimal recognition performance is achieved where the number of neurons in its hidden layer is also 65 as shown in Fig. 8. In Set#3, on the curve of Fig. 9, the recognition performance is highest with 70 hidden neurons. However, the average recognition performance among the three sets is best with 65 neurons in the hidden layer. Thus the number of neurons in the hidden layer of the MLP is finally fixed up to 65. With this, the design process is completed producing an MLP (76-65-10) for recognition of handwritten numerals on the basis of the feature set explained before. The average Recognition performance of this MLP on the test sets, as observed with 65 hidden neurons, is 96.67%.

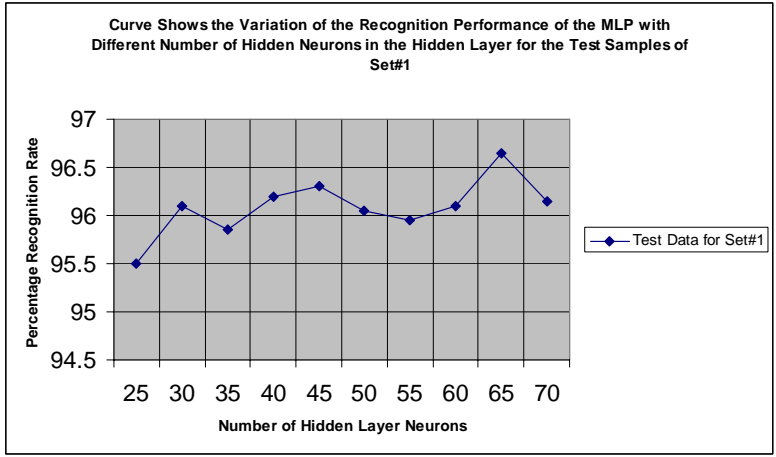

**Fig. 7.** Curves show variation of recognition performances of the MLP as the number of neurons in its hidden layer is increased in Set#1.

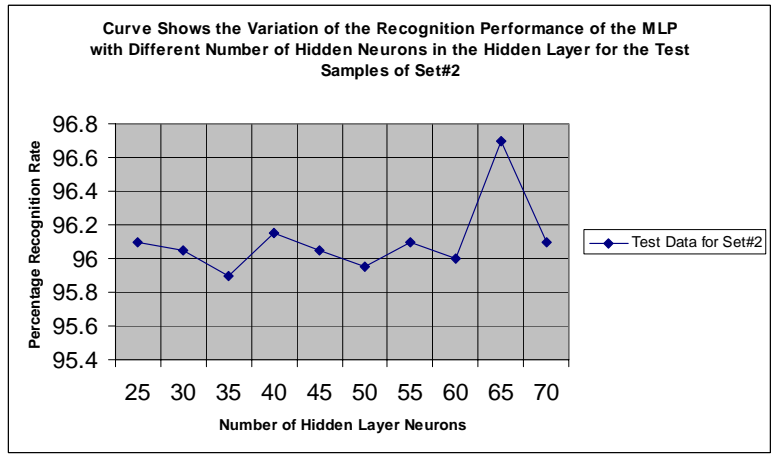

**Fig. 8.** Curves show variation of recognition performances of the MLP as the number of neurons in its hidden layer is increased in Set#2.

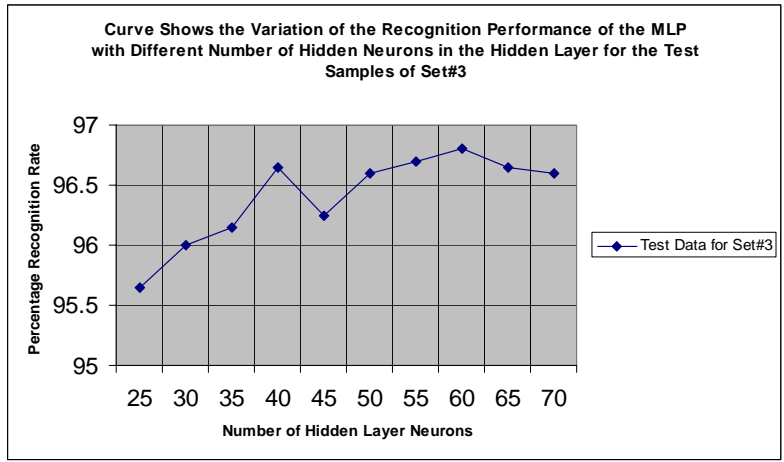

**Fig. 9.** Curves show variation of recognition performances of the MLP as the number of neurons in its hidden layer is increased in Set#3.

Recognition performances of the MLP can be further improved *firstly* by adding newer variation of handwritten numeric samples to the training set and *secondly* considering more discriminating features for digits. The work presented here can have useful application in the development of a complete OCR system for handwritten Bangla script.

## Acknowledgement

Authors are thankful to the "Center for Microprocessor Application for Training Education and Research", "Project on Storage Retrieval and Understanding of Video for Multimedia" and Computer Science & Engineering Department, Jadavpur University, for providing infrastructural facilities during progress of the work. Authors are also thankful to the CVPR Unit, ISI Kolkata, for providing the necessary dataset of handwritten Bangla Numerals. One of the authors, Mr. Subhadip Basu, is thankful to MCKV Institute of Engineering for kindly permitting him to carry on the research work.

2. P.K. Wong and C. Chan, "Off-line Handwritten Chinese Character Recognition as a Compound Bays Decision Problem", IEEE Trans. Pattern Analysis and Machine Intelligence, vol. 20,pp 1016-1023, 1998.

3   A. Amin "Off-Line Arabic Character Recognition: The State of the Art", Pattern Recognition, vol. 31, No. 5. pp. 517-530, 1998.

4. B. B. Chaudhuri and U. Pal, "A Complete Printed Bangla OCR System", Pattern Recognition, vol. 31, No. 5. pp. 531-549, 1998.

5. S. Basu, C. Chawdhuri, M. Kundu, M. Nasipuri, D. K. Basu, "A Two-pass Approach to Pattern Classification", N.R. Pal et.al. (Eds.), ICONIP, LNCS 3316, pp. 781-786, 2004.

6. K. Roy et al., "An application of the multi layer perceptron for handwritten digit recognition", CODEC 04, Jan.1-3, 2004, Kolkata.

7. Y.S. Huang , C.Y. Suen, "A method of combining multiple experts for the recognition of unconstrained handwritten numerals", IEEE Trans. PAMI, vol. 17,no.1,Jan. 1995, pp. 90-94.

8. Y. Xu and G. Nagy, "Prototype Extraction and Adaptive OCR", IEEE Trans. Pattern Analysis and Machine Intelligence, vol. 21,pp 1280-1296, 1999.

9. A.F.R. Rahman,R. Rahman,M.C. Fairhurst, "Recognition of Handwritten Bengali Characters: a Novel Multistage Approach," Pattern Recognition, vol. 35, p.p. 997-1006, 2002.

10. T. K. Bhowmik, U.Bhattacharya and Swapan K. Parui, "Recognition of Bangla Handwritten Characters Using an MLP Classifier Based on Stroke Features," in Proc. ICONIP, Kolkata, India, p.p. 814-819, 2004.
Proc. 2nd Indian International Conference on Artificial Intelligence, pp. 407-417, Dec. 2005, Pune.